\documentclass{article} %
\usepackage{colm2024_conference}

\usepackage{microtype}
\usepackage{hyperref}
\usepackage{url}
\usepackage{booktabs}
\usepackage{wrapfig,lipsum,booktabs}

\title{An Investigation on LLMs’ Visual Understanding Ability \\ Using SVG for Image-Text Bridging}

\colmfinalcopy

\author{
\textbf{{Mu Cai\thanks{Equal Contribution. Order determined by random dice rolling.} \hspace{1.5pt}$^{1}$, Zeyi Huang$^{*1}$, Yuheng Li$^{1}$, Utkarsh Ojha$^{1}$, Haohan Wang$^{2}$, Yong Jae Lee$^{1}$}}\\\\
{$^1$University of Wisconsin--Madison~~~~~~$^{2}$University of Illinois Urbana-Champaign} 
}

\usepackage{hyperref}
\usepackage{url}

\usepackage{enumitem}

\usepackage{booktabs}       %
\usepackage{amsfonts}       %
\usepackage{nicefrac}       %
\usepackage{microtype}      %
\usepackage{graphicx} 
\usepackage{geometry}
\usepackage{listings}
\usepackage{tabularx}
\usepackage{svg}

\begin{document}

\maketitle

\begin{abstract}

Large language models (LLMs) have made significant advancements in natural language understanding. However, through that enormous semantic representation that the LLM has learnt, is it somehow possible for it to understand images as well? This work investigates this question. To enable the LLM to process images, we convert them into a representation given by Scalable Vector Graphics (SVG). To study what the LLM can do with this XML-based textual description of images, we test the LLM on three broad computer vision tasks: (i) visual reasoning and question answering, (ii) image classification under distribution shift, few-shot learning, and (iii) generating new images using visual prompting. Even though we do not naturally associate LLMs with any visual understanding capabilities, our results indicate that the LLM can often do a decent job in many of these tasks, potentially opening new avenues for research into LLMs' ability to understand image data. Our code, data, and models can be found here~\url{https://github.com/mu-cai/svg-llm}.

\end{abstract}

\begin{figure}[htbp]
  \centering
  \vspace{-1.0em}
  \includegraphics[width=\linewidth]{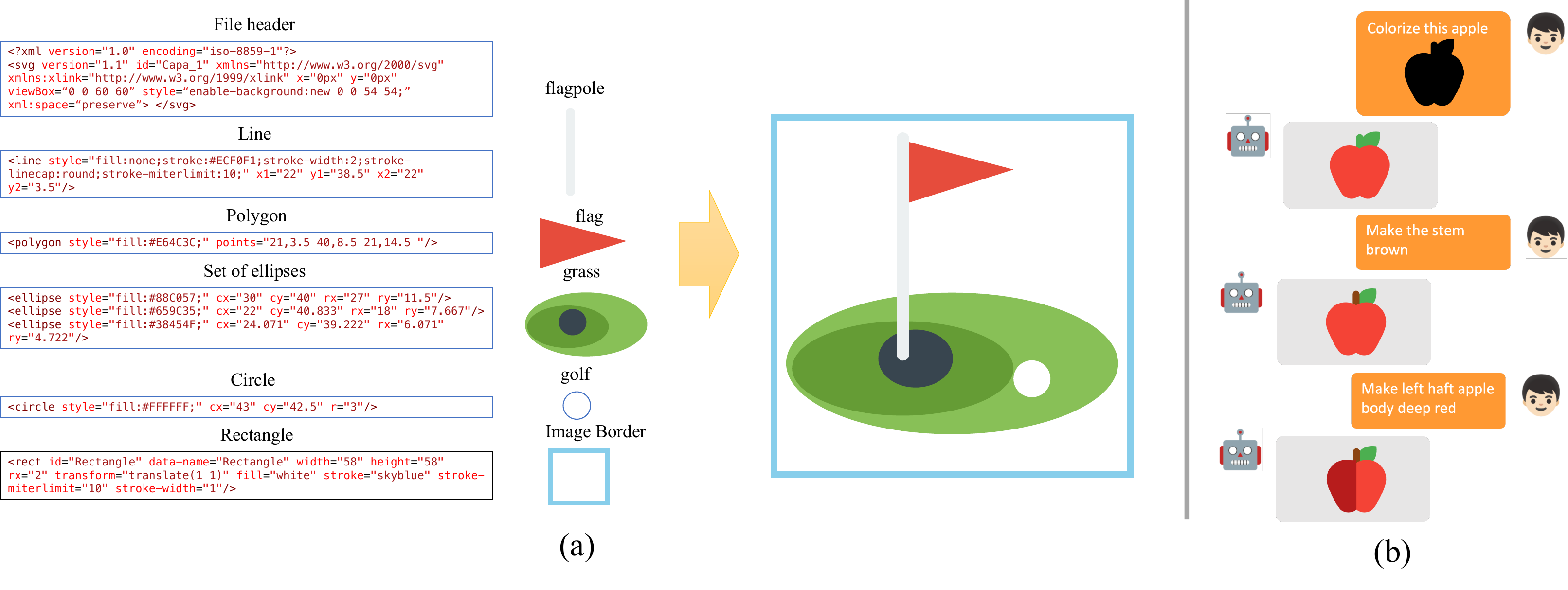}
  \vspace{-1.4em}
  \caption{(a) An SVG representation illustrating a golf course. Each geometric shape represents a distinct object. (b) LLMs can understand and generate shapes, colors, and relationships between different elements in an interactive manner.} 
  \label{fig:svg_example}
\end{figure}

\section{Introduction}
\label{sec:intro}
Large-scale data and enormous compute: the effect of these two ingredients has been on display in recent years in the significantly increased capability of machine learning systems. Models operating on the two most popular forms of data - image and text - have particularly felt that effect the most. From the side dealing with textual data, we have seen the emergence of large language models (LLMs) such as ChatGPT~\citep{chatgpt} and GPT-4~\citep{gpt4}. Similarly, on the vision side, large vision models (LVMs) have also shown impressive accomplishments~\citep{dosovitskiy2020image, liu2022convnet, dehghani2023scaling}.

When we compare these two realms, the abilities of LLMs stand out in a distinct way because of their remarkable abilities in reasoning, in-context learning, and open-ended tasks~\citep{bubeck2023sparks}. These analytical capabilities are something that the vision models, despite their significant advances, have not yet mirrored to the same depth~\citep{dehghani2023scaling, alayrac2022flamingo}.

This distinction can be attributed to the inherent nature of their respective data: LLMs thrive on the diverse and sequential structure of textual data, 
which is conducive to understanding intricate relationships and producing contextually relevant responses. In contrast, the continuous and varied nature of visual data complicates the discernment of nuanced relationships, potentially hindering the depth of analysis LVMs can achieve~\citep{bar2022visual, diwan2022winoground, lake2017building}.  Moreover, there is an ongoing debate on whether LLMs, trained on internet-scale text data, can learn world models that could lead to AGI capabilities, or that they are fundamentally limited due to their lack of grounding on physical interaction and visual perception in the real world~\citep{lecun2022path}.

The disparity between LLMs and LVMs, and the debate on the necessity of physical interaction and perceptual grounding, intrigue us to a question: \emph{Can LLMs, which have never seen visual data, understand and reason about images?}  Answering this question will bring us closer to understanding the capabilities of LLMs beyond the textual domain, their fundamental limitations, and whether they possess world models.  As such, our study takes a small but important step toward this goal.

To enable an off-the-shelf pre-trained LLM to ``read'' images, we use the Scalable Vector Graphics (SVG) \citep{ferraiolo2000scalable} representation to convert images into readable text. Unlike traditional pixel-based images, SVGs are described in XML, offering a text-based portrayal of mid-level shapes, curves, lines, and colors, as shown in Figure \ref{fig:svg_example}. The textual nature of SVG provides a data modality that LLMs excel at, acting as a bridge to apply their analytical strengths to the visual domain.  While the Sparks of AGI paper \citep{bubeck2023sparks} showed some initial qualitative results on the image understanding capabilities of LLMs using a similar idea, we provide a deeper, comprehensive study that includes both qualitative and quantitative analyses on a variety of visual understanding and reasoning tasks.

Specifically, we evaluate whether an LLM can perform both discriminative and generative visual understanding tasks. For discriminative tasks, we study their capabilities in visual reasoning, visual question answering, few-shot learning for image classification, and their robustness to distribution shift in visual data.  Surprisingly, despite never having seen dense visual data, LLMs perform much better than chance and are often robust to distribution shifts.  For generative tasks, we study LLMs' image generation ability through the task of visual prompting. We find that LLMs can identify transformations related to color, shape, style, and content from provided SVG examples, generating credible output that is consistent with those examples.

\section{Related Work}
\label{sec:related}

\subsection{Leveraging LLMs for Visual Tasks}

Upon observing the powerful reasoning capabilities of LLM, researchers began to harness its potential for visual tasks. Presently, there are three primary works to utilizing LLM for these purposes.  The first work involves using LLM to produce textual guidelines. Vision models then rely on these instructions to execute a range of visual tasks. Examples include Visual ChatGPT~\citep{wu2023visual}, visual programming as seen in~\cite{gupta2023visual}, and ViperGPT~\citep{suris2023vipergpt}.
The second work, as illustrated by LLaVA~\citep{liu2023llava} and MiniGPT4~\citep{zhu2023minigpt}, incorporates the pretrained vision encoder model, along with a trainable projector. This allows for feeding visual features directly to LLMs, demonstrating remarkable reasoning abilities. VisionLLM~\citep{wang2023visionllm} presents bounding boxes and segmentation masks in text format (as polygons), enabling LLMs to address more intricate perception challenges.
The third work seeks to represent images directly in a text-based format, bypassing the use of visual encoders. The goal here is to allow LLMs to interpret these text-based representations. For instance, LIFT~\citep{dinh2022lift} represents images using their raw pixel values in textual form and then fine-tunes the language model on them for visual tasks. Another study~\citep{bubeck2023sparks} explores image generation by expressing the image in text formats, like TiKZ or SVG.

\textbf{Our novelty:} 
Our study aligns with the third direction. Unlike LIFT~\citep{dinh2022lift}, we represent images using SVG, a format that inherently encodes more structural information than raw pixel values. This could enable LLMs to better grasp intricate relationships and yield contextually relevant responses. Distinct from the investigation presented in~\cite{bubeck2023sparks}, we conduct a comprehensive study of how LLMs process images via textual representations, including both discriminative and generative tasks.

\subsection{Scalable Vector Graphics}

Vector graphics describe images as collections of parameterized shape primitives such as polygons, circles, and rectangles,  rather than a regular raster grid of pixel values~\citep{peng2004roles}. Primitives are usually characterized by a set of coordinates delineating their contour and the associated color. This leads to a compact and infinitely scalable representation where the appearance can be easily modified by adjusting stroke or color parameters. Consequently, vector graphics are the preferred choice among graphic artists and designers, as images maintain their sharpness regardless of the zoom level. Encapsulated PostScript (EPS) and Scalable Vector Graphics (SVG) are two notable vector-based formats~\citep{ferraiolo2000scalable}.
SVG format stores images as XML-based text files that define geometrical objects and their properties~\citep{ferraiolo2000scalable}, as shown in Figure~\ref{fig:svg_example}. This enables easy editing, manipulation, and embedding, which makes SVG particularly versatile for web applications and graphic design tasks~\citep{badros2001constraint}. EPS is another vector format for high-quality graphics that can be resized without losing quality~\citep{gruber2008vienna}. 
In this paper, we systematically evaluate LLMs’ abilities to generate and understand an assortment of visual concepts.

\subsection{Large Language Models}

Large Language Models (LLMs) have attracted much attention in recent years due to their remarkable performance across numerous natural language processing tasks. GPT-3~\citep{NEURIPS2020_1457c0d6}, developed by OpenAI, is a prime example of this category, boasting an immense scale of 175 billion parameters and human-like text generation capabilities. In a similar vein, BERT~\citep{devlin-etal-2019-bert} (Bidirectional Encoder Representations from Transformers), introduced by Google, takes advantage of the transformer architecture and has substantially enhanced the state-of-the-art across various tasks by learning deep bidirectional representations. ChatGPT~\citep{chatgpt}, another noteworthy model, is a GPT variant specifically designed for human-like conversational abilities. The most recent iteration, GPT-4~\citep{gpt4}, succeeds GPT-3~\citep{brown2020language} and carries on the LLM advancements in terms of scale and performance. These models lay the groundwork for our research, enabling us to investigate their potential in more complex tasks such as image understanding and generation. Our work illustrates the applicability of LLMs to SVG-based image understanding and generation, paving the way for novel applications and research directions in the visual domain.

\vspace{-1em}
\section{Tasks and Experimental Results}

\label{sec:Experiments}

There are many different kinds of visual understanding problems, and often times, models need to have different kinds of abilities to solve them. In this section, we wish to investigate if LLMs indeed have those required abilities. However, for them to solve a visual understanding problem, there needs to be a way for them to \emph{see} an image. SVG can be that bridge, where an image is converted into a structured XML code (see Fig.~\ref{fig:svg_example} for an example). And like any other code, LLMs can potentially \emph{read} the code and perform some task of interest. To test what is possible using this form of image representation, we consider three broad categories of visual understanding tasks. 

In Sec.~\ref{sec:reasoning}, we first study the problem of visual reasoning, where the model is asked certain kinds of questions about the contents of an input image (e.g., \textit{How many objects have the same shape as the red object?}).
Next, in Sec~\ref{sec:robustness}, we study how LLMs fare in image classification tasks; especially under non-trivial settings like distribution shifts and few-shot learning.
Then, in Sec~\ref{sec:robustness}, we demonstrate SVGs' robustness to permutations. 
Furthermore, in Sec~\ref{sec:QA}, we explore the understanding of SVG by LLMs in visual question answering tasks.
After studying their abilities in discriminative tasks, in Sec~\ref{sec:Generation}, we then test the generative modeling capabilities of LLMs through the task of visual prompting, which asks the model to generate a new image following the pattern depicted by three images - A : B, C : ?. Finally, due to page constraints, we delve into additional generative tasks such as style and content extrapolation, generation with human feedback, and referring segmentation in the supplementary materials.

Unless otherwise mentioned, all the experiments use GPT-4~\cite{gpt4} as the LLM model. The common theme across all these different experiments is that the LLM processes the images converted in the form of SVG. %

\subsection{Visual reasoning}

\label{sec:reasoning}

\begin{figure*}[t]
\centering
\includegraphics[width=\textwidth]{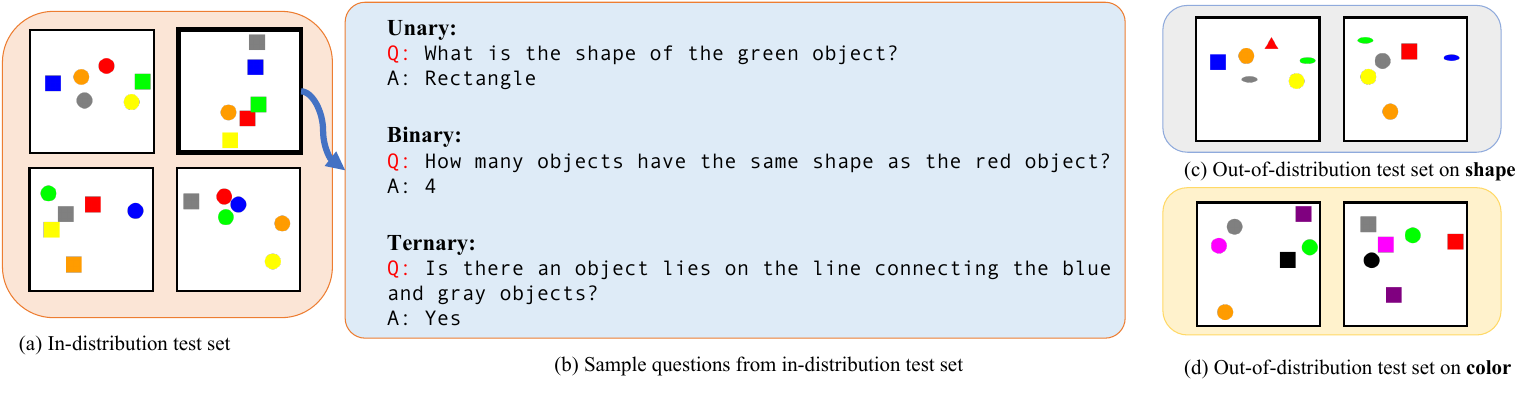}
\caption{Illustration of In-Distribution and Out-of-Distribution Test Sets:
(a) Images from the in-distribution test set, showcasing random sampling of object color, shape, and location.
(b) Accompanying each image are questions assessing unary, binary, and ternary reasoning capabilities.
(c) Expansion of shape variety to include ellipses alongside rectangles and circles.
(d) Introduction of additional colors, including magenta, black, and purple, to the sampling palette. }
\label{fig:clevr}
\end{figure*}

\begin{table*}[t]
\caption{Visual reasoning performance of Sort-of-Clevr dataset under in-distribution test set and out-of-distribution test set with shape and color distribution shift.}
\label{table:sort-of-clevr-shift}
\scriptsize 
\centering
\setlength\tabcolsep{4pt}
\begin{tabular}{lccccccccccc}
\toprule 
Distribution shift & \multicolumn{4}{c}{i.i.d.} & \multicolumn{3}{c}{o.o.d. Shape} & \multicolumn{3}{c}{o.o.d. Color} \\
\cmidrule(lr){2-5} \cmidrule(lr){6-8} \cmidrule(lr){9-11}
Question type & GPT4-brief & GPT4-CoT & CNN & Rel. Net. & GPT4-CoT & CNN & Rel. Net. & GPT4-CoT & CNN & Rel. Net. \\
\midrule
Image format  & SVG & SVG & PNG & PNG & SVG & PNG & PNG & SVG & PNG & PNG \\
\midrule
Unary   & 0.50 & 0.90  & 0.65 & 0.89  & 0.95 & 0.58 & 0.82 & 0.95 & 0.56 & 0.83 \\
Binary  & 0.90  & 0.95 & 0.75 & 0.80  & 0.95 & 0.36 & 0.44 & 0.95 & 0.57 & 0.66 \\
Ternary  & 0.10 & 0.88  & 0.55 & 0.55  & 0.63 & 0.52 & 0.56 & 0.71 & 0.47 & 0.54 \\
\midrule
 Average   & 0.50 & 0.89  & 0.65 & 0.75 & 0.84 & 0.49 & 0.61 & 0.87 & 0.53 & 0.67 \\
\bottomrule
\end{tabular}
\vspace{-1em}
\end{table*}

With all the successes in the traditional perception tasks like image classification/segmentation~\cite{liu2022convnet,kirillov2023segment}, visual reasoning still remains a pivotal challenge for many modern computer vision systems~\cite{zhang2019raven, barrett2018measuring, santoro2017simple}. It typically refers to the ability of a model to answer questions about different constituents of an image. To better understand what entails that task, we first discuss the dataset we use for this experiment.

\paragraph{Dataset:} We use the Sort-of-Clevr dataset~\cite{santoro2017simple}. As shown in Figure~\ref{fig:clevr}, each image in the dataset is composed of 6 objects, each with a unique color, having a randomly chosen shape between rectangle and circle. For each such image, the dataset contains many \{question, answer\} pairs. There are three categories of questions crafted to test three levels of reasoning capabilities: unary, binary, and ternary relationships. An example question testing binary reasoning ability is - \textit{What is the shape of the object closest to the gray object?} Please refer to Fig.~\ref{fig:clevr} for more examples of the questions. Each question is, ultimately, a classification problem. The methods are evaluated based on their top-1 accuracy on questions from the test set of the dataset. Please refer to appendix for the details about the test sets used for evaluation.

\paragraph{Methods:} Broadly speaking, we show the results on two kinds of methods - (i) LLMs which are \emph{not} trained for this task, and are only prompted during inference, and (ii) methods which are trained for this task on the training set of Sort-of-Clevr dataset. Among the first category, we analyze (a) GPT4-brief~\citep{gpt4}, (b) GPT-CoT (Chain of thought), and (c)  GPT-4V~\cite{gpt4} and open source multimodal models including LLaVA~\citep{liu2023llava}, InstructBLIP~\cite{instructblip}, BLIP2~\cite{li2023blip}, mPLUG\_owl~\cite{ye2023mplug}, and MiniGPT4~\cite{zhu2023minigpt}. Among the second category, we study two methods: (a) CNN+MLP and (b) Relation Networks~\citep{santoro2017simple}. The reason we choose to evaluate on this category of trained models is to have an idea of what the upper bound could be; i.e., how difficult the task really is. To evaluate any of the LLMs during inference, we transform the original images into geometric primitives into their SVG format. Then we query the LLM using the following prompt: e.g., "Given the following SVG image \texttt{<svg>...</svg>}, what is the shape of the red object?" The difference between GPT4-brief and GPT4-CoT is the way we ask the final question: in GPT4-brief, as the name suggests, our final question asks the model to provide the answer briefly, whereas in GPT4-CoT, we explicitly ask the model to break down its reasoning before arriving at the answer (please see the appendix for the exact prompt used to elicit this behavior). By querying an LLM, we obtain answers which are then summarized post human evaluation to determine the final accuracy. Due to costs associated with probing GPT4 models, our evaluation is restricted to 120 examples.

Table~\ref{table:sort-of-clevr-shift} shows the top-1 accuracy of different methods (left: zero-shot inference of LLMs, right: methods trained for the task). Note that the results of multimodal models such as LLaVA are shown in the appendix. When looking at the two LLM models that process images in the SVG format, GPT4-brief, and GPT-CoT, we can first see that their performance is much higher than chance (many questions in the test set have 5-6 correct answers, thereby reducing the chance performance accuracy; please see the appendix). Furthermore, the performance of GPT-CoT even surpasses the performance of a model explicitly trained for this task. If we take a step back and think once more about the nature of SVG representation (Fig.~\ref{fig:svg_example}), the best-case scenario might be when images from the Sort-of-Clevr dataset have the locations of certain shapes embedded in their XML code. But even if such a nicely structured code is often available to the LLM, to properly be able to reason about questions like the ones described in Fig.~\ref{fig:clevr}, the LLM needs to precisely perform many mathematical relational operations. From that perspective, the results depict that LLMs might possess much complex models already.

\textbf{Distribution shift evaluation:} Furthermore, to study an even more difficult version of the problem, we test the performance of models under distribution shifts. Specifically, we evaluate under both shape and color distribution shifts. As for the color, we replace 3 colors from the original 6 colors with the new colors. For shape, we randomly enlarge the options to further include the ellipse and triangle, as shown in Figure~\ref{fig:clevr} (c) (d). As a result,  each object can sample the shape uniformly from the 4 choices. Importantly, we make sure that all visual reasoning questions can be answered using the original one-hot choices for vision models like CNN-MLP and relation networks. 

As shown in Table~\ref{table:sort-of-clevr-shift}, the LLM model (GPT4-CoT) using the SVG format to process images does not suffer much by any of the newly added complications in the test images (e.g., more shapes added under the shape distribution shift), maintaining its ability to perform the reasoning tasks. Here is what this means in simple words, as understood through an example: if in the original image (before distribution shift), there was a red circle immediately to the left of a blue rectangle, even after introducing other shapes (e.g., triangles), the LLM can still detect the red circle to the left of that rectangle. This is not trivial, because the models which were explicitly trained on the Sort-of-Clevr dataset do suffer a non-trivial loss in performance; both in color and shape distribution shift. Overall, these results indicate that the internal model used by the LLM is surprisingly effective at tasks that we wouldn't have naturally thought of it being good at.

\begin{table*}[t]
\caption{Image classification results with both vision and language models. We utilize the Mini-MNIST dataset to evaluate GPT4's ability to understand SVG through both zero-shot and one-shot in-context learning. To evaluate the model's robustness against distribution shift, vision model ConvNeXt and language model Vicuna are finetuned on the MNIST training set and evaluated on the MNIST test set, CMNIST-A, and CMNIST-B respectively. ICL denotes in-context learning. Additionally, Vicuna's performance with PNG inputs is compared.}
\label{table:classification}  
\small  
\centering  
\begin{tabular}{l|c|cc|c|c}  
\toprule  
Method  & ConvNeXt(fine-tuning)  & \multicolumn{2}{c}{Vicuna(fine-tuning)} & GPT4(Zero-Shot) & GPT4(One-Shot ICL) \\  
\midrule  
Image Format & PNG & SVG & PNG & SVG & SVG\\  
\midrule  
MNIST    &  99.5\% &      99.1\%  &  99.4\% &  20\% &  24\%\\  
\midrule  
CMNIST-(A) & 79.5\% &  95.7\% & 42.9\% & 16\% & 19\%  \\  
CMNIST-(B)  & 32.6\%  &  92.9\%  & 24.8\% & 13\%  & 20\% \\  
\bottomrule  
\end{tabular}  
\vspace{-0.8em}
\end{table*}

\begin{figure}[t]
  \centering
  \includegraphics[width=0.7\linewidth]{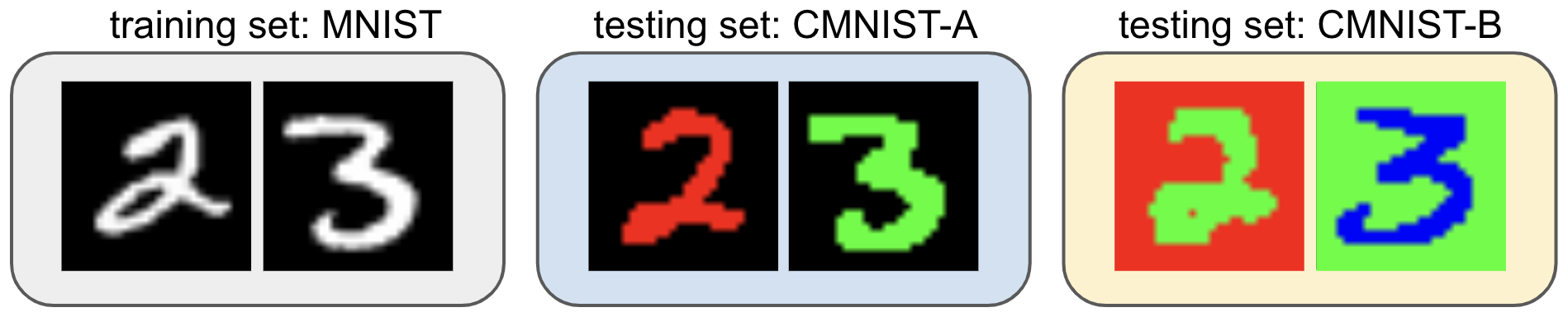}
\caption{Out-of-Distribution generalization tasks, where we wish to train a model on the standard gray-scale MNIST and test on variants of colored MNIST.}
  \label{fig:svg_cmnist}
  \vspace{-1.2em}
\end{figure}

\subsection{Out-of-distribution Generalization}
\label{sec:robustness}

To DNNs, innocuous transformations can completely change predictions. This has been reported in various cases
such as shifting the image by a few pixels~\citep{azulay2018deep}, adding a bit of random noise~\citep{hendrycks2019benchmarking} or changing the background, color, or texture~\citep{he2021towards,arjovsky2019invariant,geirhos2018imagenet} while keeping the shape intact. In this section, we aim to investigate if representing images as SVG could mitigate these issues and result in a model focusing more on recognizing the target shapes.

\textbf{Datasets}: We construct two variants of the Colored-MNIST dataset to assess model robustness against color and background variations. The first version, termed Colored-MNIST-A, assigns a color of either red or green to the foreground, with each color being selected randomly at an equal likelihood of 50\%. In the more challenging second version, dubbed Colored-MNIST-B, both the background and foreground are selected from a color palette that includes black, white, red, blue, and green. The background and foreground colors are always distinct, yielding 20 unique color combinations. Visualization of these Colored-MNIST datasets can be viewed in Figure~\ref{fig:svg_cmnist}. Furthermore, we utilize the curve tracing algorithm to convert MNIST images into the SVG format. More details can be found in the appendix.

\textbf{Task and experimental setting}: In the first setup, we fine-tune the ImageNet pre-trained vision model ConvNeXt~\citep{liu2022convnet} using PNG images and the pre-trained language model Vicuna~\cite{Vicuna} using both PNG images and SVG-converted images on MNIST. Subsequent testing is carried out on both Colored-MNIST variants (A) and (B). This setup seeks to examine whether the model can prioritize shape over other features for its predictions. In the second setup, our objective is to explore the potential of harnessing the potent in-context capabilities of Large Language Models (LLMs) to enhance image classification using SVG. So, we employ GPT-4~\citep{gpt4} to conduct both zero-shot and in-context learning on MNIST variants. More details on ConvNeXt and Vicuna fine-tuning, prompting for in context learning can be found in the appendix.

\textbf{Results and discussion}: 
In Table~\ref{table:classification}, 
fine-tuning Vicuna with SVG representation shows promising results on the CMNIST-A and CMNIST-B benchmarks, achieving accuracies of 95.7\% and 92.9\% respectively. This suggests some level of robustness against color and background perturbations. On the other hand, ConvNeXt and Vicuna under PNG format are more susceptible to these perturbations, with a noticeable decline in performance on both benchmarks compared to i.i.d results. We hypothesize that SVG might offer a representation more biased towards shape, given its explicit textual encoding of object shapes, allowing for disentanglement of shape from color information.
Further, as illustrated in Table~\ref{table:classification}, there's a notable 4\% accuracy boost when using a single in-context sample, as compared to a zero-shot classification approach. 
This demonstrates the capability of LLM to grasp visual concepts contextually.

\begin{table*}[t]
\caption{Synthetic data study results (color-aware mIOU) on the six tasks~\citep{bar2022visual}. GPT4 is better able to understand and reason about different transformations using SVG representation.}
\label{table:generation}
\small
\centering
\begin{tabular}{lcccccc}
\toprule
Method & Color & Shape & Size & Color Shape & Color Size & Shape Size \\
\midrule
VQGAN~\citep{esser2021taming} & 7.0 & 19.1 & 16.2 & 7.4 & 2.2 & 18.4  \\
BEiT~\citep{bao2022beit} & 40.9 & 31.4 & 7.1 & 33.1 & 21.2 & 13.0  \\
MAE~\citep{he2022masked} & 70.2 & 44.0 & 34.7 & 19.3 & 19.0 & 46.0  \\
MAE-VQGAN~\citep{bar2022visual} & 40.4 & 46.5 & 42.0 & 20.4 & 18.3 & 40.3  \\
\midrule
SVG with GPT4 & 100.0 & 92.6 & 100.0 & 92.6 & 100.0  & 86.5 \\
\bottomrule
\end{tabular}
\end{table*}

\subsection{Visual Question Answering}
\label{sec:QA}

To quantitatively and comprehensively how well large language models can understand SVGs, we collect a large-scale SVG question-answering dataset with 2318 QA pairs under three hierarchical levels including color, category, and usage. 

\begin{figure}[htbp]
  \centering
  \includegraphics[width=\linewidth]{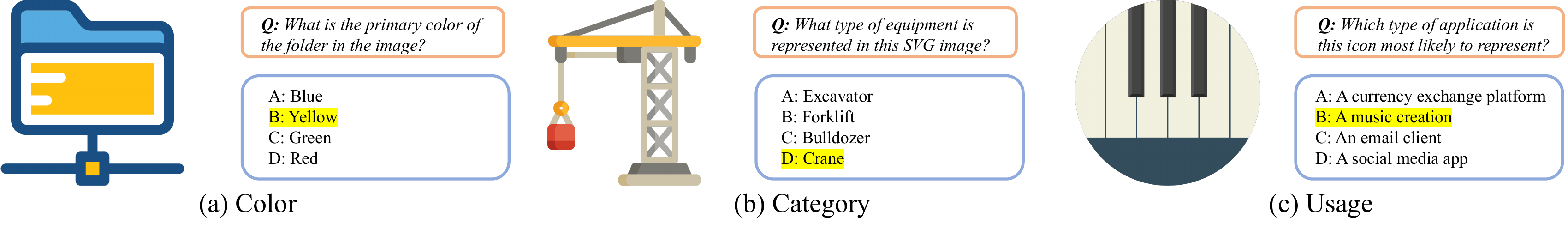}
  \caption{The samples SVG question-answering dataset, which belongs to the Color, Category, and Usage task, respectively. }
  \vspace{-0.8em}
  \label{fig:svg_qa}
\end{figure}

\textbf{Datasets}: We collect the dataset following 2 steps. First, we leverage GPT-4V to automatically generate the question, four options along with a solution. This process involves feeding the rasterized SVG image, task instruction, and two in-context examples into GPT-4V. The detailed prompts are shown in the appendix. After this step, we collect 1000 questions for each type. Next, we ask human annotators to filter out any QAs that contain either wrong or unreasonable questions/options/answers, where we finally get 671, 869, and 778 high-quality SVG QAs.  The examples are shown in Figure~\ref{fig:svg_qa}.

\textbf{Task and experimental setting}: In this task, LLMs are tasked with answering various questions regarding the attributes of objects, including their color, category, and usage. We query representative LLMs, including GPT-3.5~\cite{chatgpt}, GPT-4~\cite{gpt4}, and the recent SoTA open-sourced LLM Mixtral-8x7B~\cite{jiang2024mixtral}. After getting their responses, we leverage GPT-4 to rate whether the LLM prediction is correct again using in-context examples.

\begin{table}
    \centering
    \caption{Results of GPT-4, GPT3.5, and Mixtral under each type of question task: Color, Category, and Usage. The results showcase that current LLMs have a certain level of SVG understanding capabilities in object's color, category, and usage.}
    \scalebox{1}{
    \begin{tabular}{l|ccc}
    \toprule
     & GPT-3.5 & Mixtral-8x7B & GPT-4 \\\hline
    Color & 32.7 & 49.8 & 55.7 \\
    Category & 25.7 & 27.6 & 41.7 \\
    Usage & 37.5 & 38.6 & 49.1 \\
    Average & 32.0 & 38.7 & 48.8 \\
    \bottomrule
    \end{tabular}}
    \label{tab:benchmark}
\end{table}

\textbf{Results and discussion}: As shown in Table~\ref{tab:benchmark}, GPT-4 shows a nontrivial average accuracy of 48. 8\%, which means that GPT-4 can understand both semantics and low-level information of SVGs to a reasonable extent. For less capable models like GPT-3.5, the performance is less desirable but still higher than the random chance (25\%). Surprisingly, open-sourced LLM Mixtral-8x7B  shows stronger SVG understanding capability than GPT-3.5 across all aspects, specifically for color understanding. Our benchmark demonstrates that both open-sourced and closed-sourced LLMs have certain level of the SVG understanding ability, where they excel in color recognition, while showing lower performance in category (semantic) understanding.

\subsection{Visual Prompting}
\label{sec:Generation}

\begin{figure}[t]
\centering
\includegraphics[width=\textwidth]{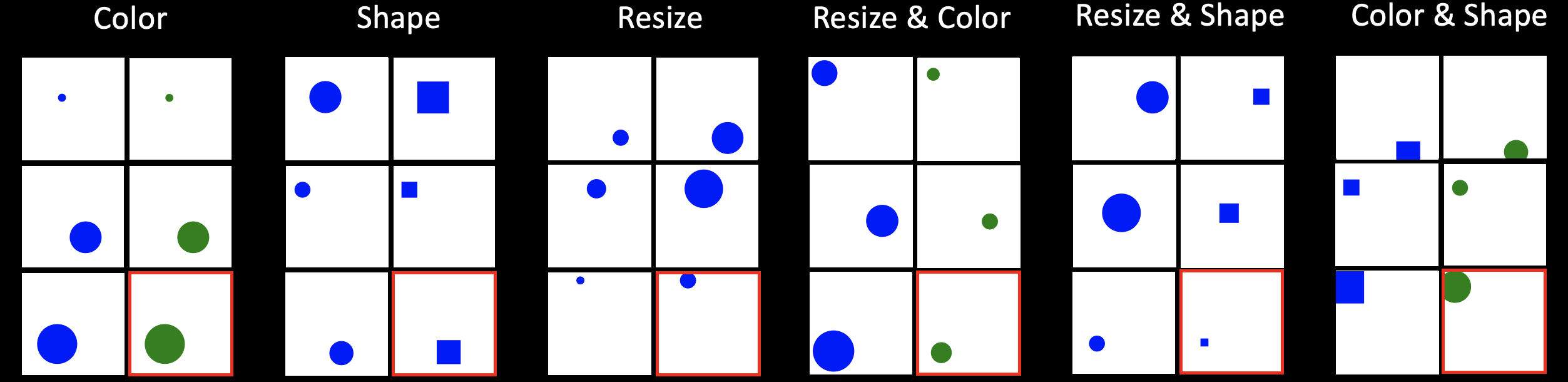}
\caption{Understanding SVG transformations through the lens of GPT-4.  GPT4 clearly understands simple shape, color, and size transformations by analyzing the SVG code without any pixel-level information. The generation results of our method are annotated with a red square. }
\label{fig:synthetic}
\end{figure}

The previous sections discussed the emergent abilities of LLMs in discriminative tasks. Now, we turn our attention towards the generative side to see if LLMs can understand and generate logically coherent images as well. We consider the task of visual prompting, where, given a series of images, the goal is to learn the transformation and \emph{fill in} the remaining spot with an appropriate image.

\paragraph{Dataset and evaluation.}
We follow~\cite{bar2022visual} to create a set of three tasks of filling in the remaining spot (Fig.~\ref{fig:synthetic}), and three of their combinations, and evaluate each model on 100 examples per task. Every pair in our example set includes an SVG showcasing a colored shape along with a corresponding SVG with specific transformations. The transformations consist of color, size, or a combination of these aspects.  We explain the descriptions of each task in the appendix.
For evaluation purposes, we adopt the method from~\cite{bar2022visual}, for measuring and reporting the color-aware mean Intersection over Union (mIOU). Given two example pairs and a query SVG, we structure the text prompt in the same fashion for all tasks. The prompt is designed to figure out the common transformation in the two examples first and then transform that query into the corresponding key SVG code. We include the prompt details in the appendix.

\paragraph{Qualitative and quantitative results.} The results are presented in Table~\ref{table:generation}. See Figure~\ref{fig:synthetic} for our generated results. GPT4 clearly understands simple shape, color, and size transformations by analyzing the SVG code without any pixel-level information.

\section{Conclusion}

This paper studied whether LLMs can ``see'' and understand images through the Scalable Vector Graphics (SVG) format. By converting raster images into SVG representations, we provided LLMs with a textual representation of visual data, enabling them to engage in a variety of computer vision tasks. 
Our exploration across three major domains—visual reasoning and question answering, image classification in the face of distribution shifts and with limited examples, and the generation of new images through visual prompting. Despite their primary design for natural language understanding, LLMs demonstrate a promising capacity to understand and generate visual content when it is represented in an SVG format. Additionally, we have collected a SVG question-answering dataset for future research aimed at assessing the visual understanding capabilities of LLMs.
We believe that our work provides an exploratory step for understanding the capabilities of LLMs beyond the textual domain, whether they possess world models, and what their potential is.  %

\section{Limitations}

While the structured nature of the SVG representation does make them effective for certain visual tasks, that format also abstracts away the finer details of the natural images (e.g., the furry texture of a dog). To understand the negative effect of this abstraction, we consider the image classification task; specifically, 10-way classification on Imagenette~\cite{imagenette} dataset. We convert the images from the dataset into their corresponding SVG format and use the Vicuna-7B model to get the predicted class. This prompting method gives a top-1 accuracy of 68.14\%. We also train a ResNet50 model on the original images, which achieves an accuracy of 90.62\%. These results show that the loss of information during the SVG conversion can lead to a loss of performance for certain tasks.

\bibliography{custom}
\bibliographystyle{colm2024_conference}

\appendix
\section*{Appendix}

\section{Qualatative Chat Results}

\begin{figure*}[htbp]
  \centering
  \includegraphics[width=1\linewidth]{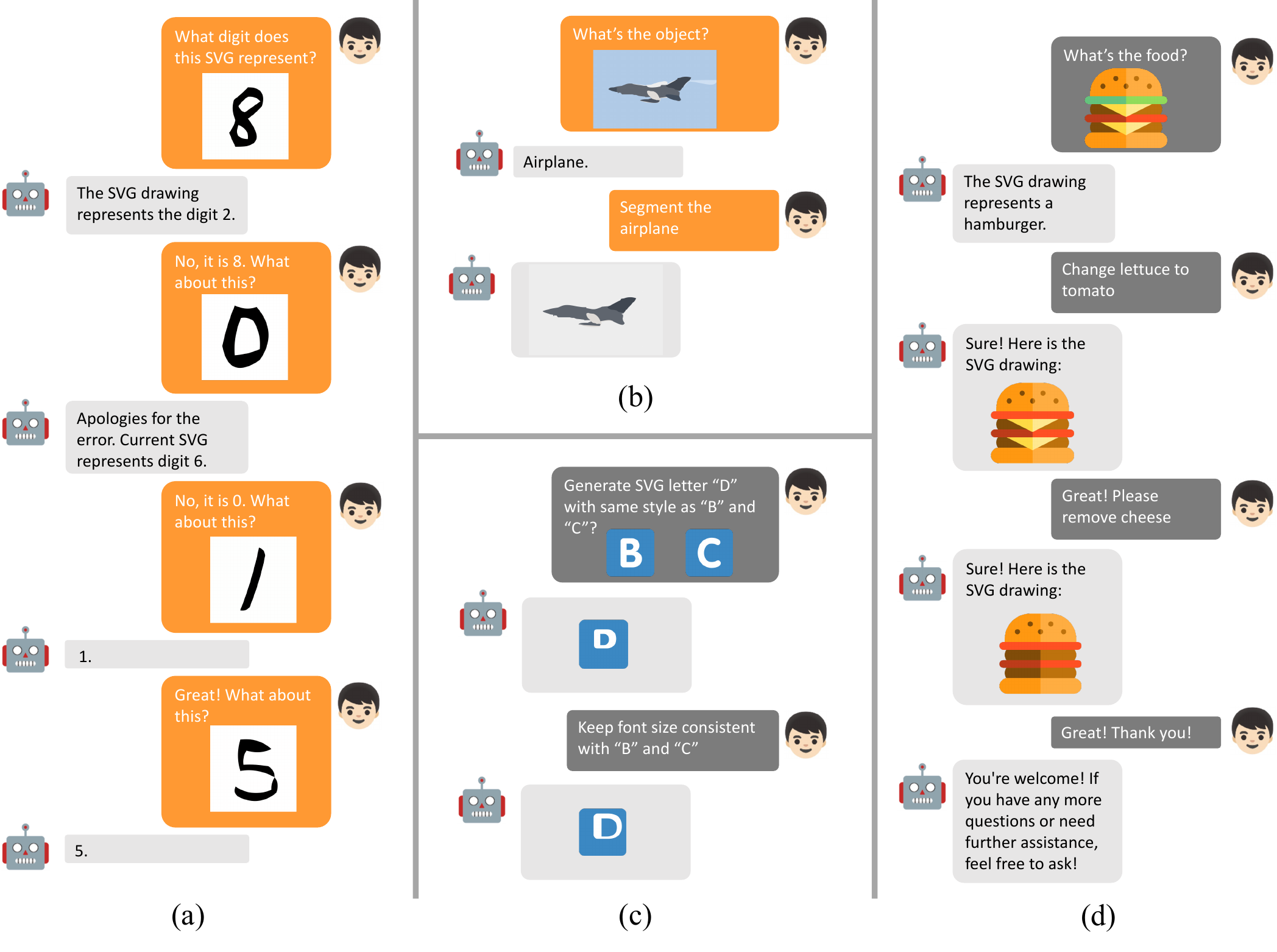}
  \caption{In-context learning and image generation capabilities of SVG with LLMs. (a) With human feedback, LLM gradually performs better on digit classification. (b) LLM powers SVG with the capability of image recognition and referring segmentation. (c) With human feedback, the content generation performance becomes better. (d) LLM can recognize and manipulate specific parts of the hamburger, such as removing or replacing them.}
  \label{fig:chat_example}
\end{figure*}

\begin{figure}[htbp]
\centering
\includegraphics[width=\textwidth]{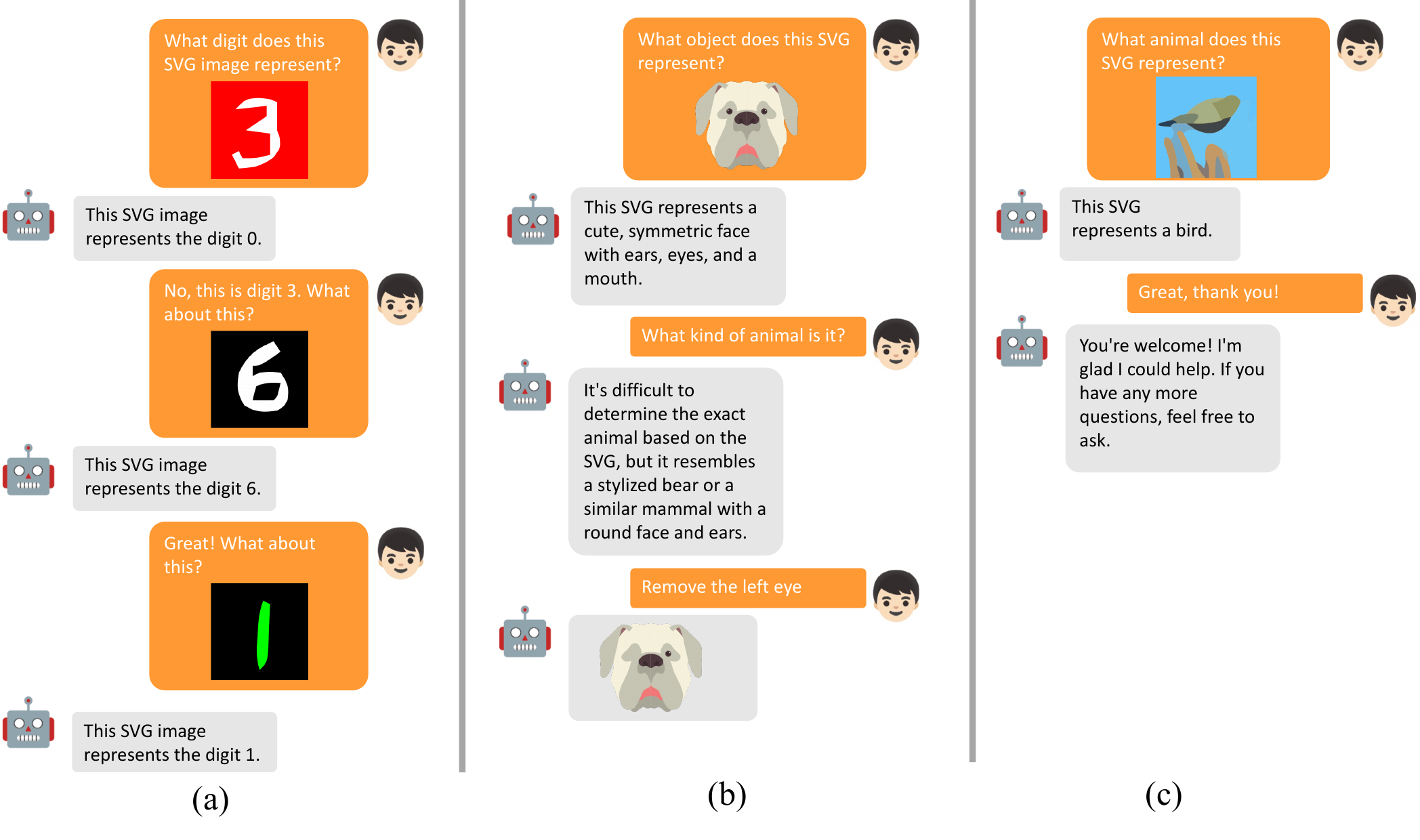}
\vspace{-2em}
\caption{More qualitative results of chat-based image recognition and manipulation. (a) In-context digit recognition in Colored-MNIST-(B). (b) GPT can explain and manipulate the dog SVG image.  (c) GPT4 can also recognize the bird from a CIFAR-10 example. }
\label{fig:chat2}
\end{figure}

\paragraph{Image Recognition and Manipulation. }
In this section, we provide  examples for chat-based image recognition and manipulation using GPT4~\citep{gpt4}. The qualitative results are shown in Figure~\ref{fig:chat_example} and~\ref{fig:chat2}: (a) SVG representation empowers robust in-context digit recognition capability given different background and foreground colors, (b) GPT4 can recognize and depict the details of a dog with the prompt: \texttt{"a stylized bear or a
similar mammal with a
round face and ears."} Furthermore, GPT-4 can identify the location of the dog's left eye and remove it.  (c) GPT4 is capable of recognizing a natural image from the CIFAR-10 dataset.

\paragraph{Referring Segmentation}

The objective of the task is to  label pixels in an image or video that correspond to an object instance referred by a linguistic expression. SVG representation has two advantages. First, language instruction is naturally embedded within the prompt, thus a separate design of the image segmentation model is not needed. Second, a large corpus of text and programming languages including XML are seen during pretraining, benefiting the vision-language understanding ability. 

SVG is typically composed of several colored polygons, where each of them can correspond to a part of the object. Therefore, we can use the referring segmentation instructions to guide the LLM in finding the corresponding SVG code.  Shown in Figure~\ref{fig:chat_example}~(b) and (d), LLM can localize the object decently well. In (b), the majority of the airplane is selected as foreground, while in (d), not only is the lettuce recognized, but also the two pieces of cheese are localized and subsequently removed.

\begin{figure}[htbp]
\centering
\includegraphics[width=\textwidth]{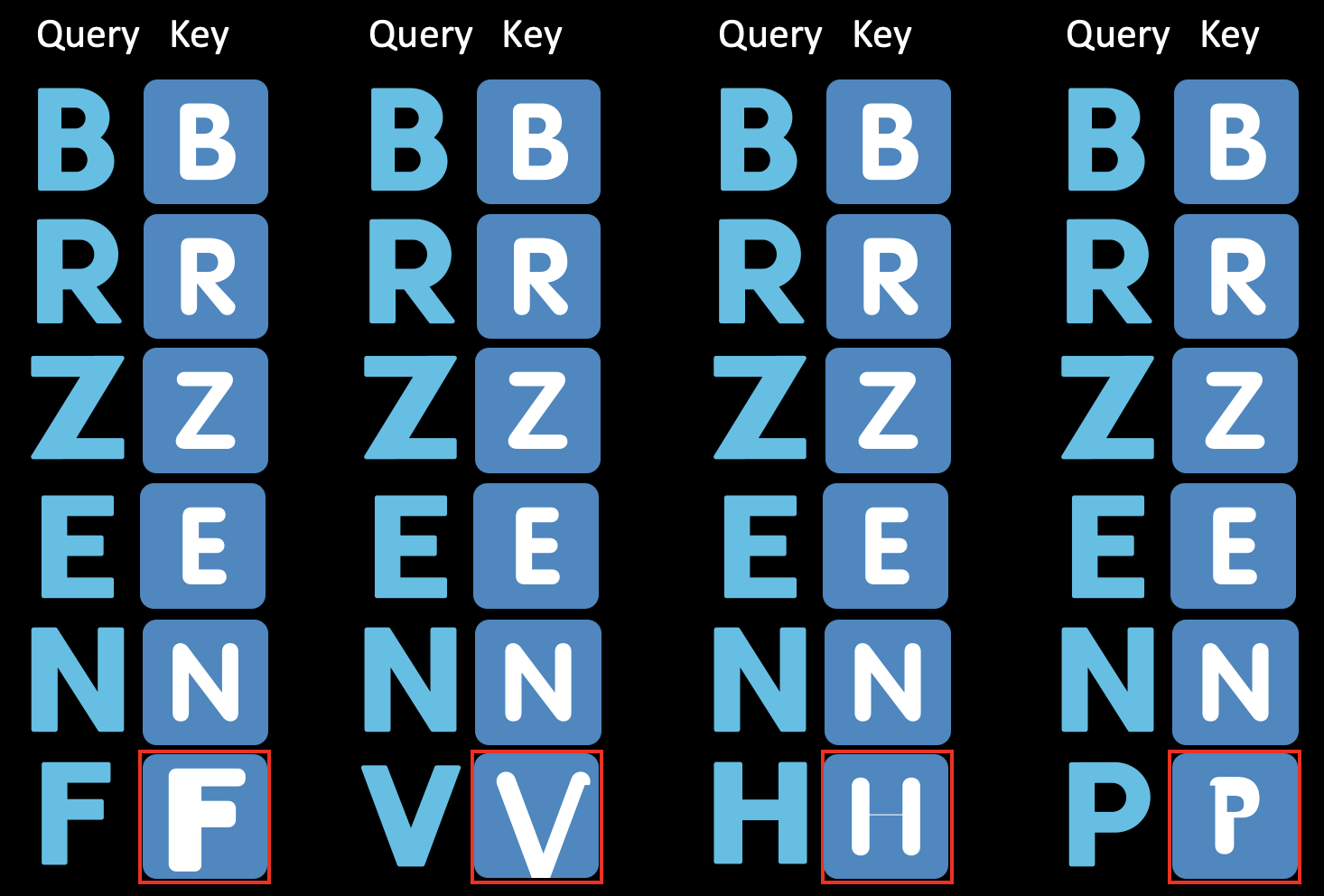}
\caption{More qualitative results of style extrapolation.  The generation results of our method are annotated with a red square.}
\label{fig:synthetic2}
\end{figure}

\section{Robustness to Permutations}
\label{sec:Permutations}

Next, to evaluate the robustness of LLMs against variations in SVG data, we conduct three distinct experiments, where we: (i) shuffle the order of paths, (ii) randomize path coordinate replacement, and (iii) randomize string replacement. Each experiment is designed to mimic real scenarios of imperfections in SVG data, providing insights into using SVG with LLMs under challenging conditions.

\textbf{Path Shuffle Experiment.} 
SVG data consists of multiple path elements, each representing an object or line in the image. In this experiment, we test the model's ability to interpret hand-drawn SVG data from the MNIST dataset when the sequence of path elements is shuffled. Note that, by the very nature of SVG data, this alteration will not change the final image; and hence, it is important to test if the LLM can remain invariant to this change, even though we have not explicitly provided it with this knowledge. The goal is to assess the model's ability to correctly interpret the digit, regardless of the order of path elements. The results in Table~\ref{tab:shuffle} indicate that LLMs are robust to path shuffle at test time, maintaining a comparable performance regardless of the path ordering.

\begin{table}[h]  
\centering  
\small
\vspace{-1em}
\caption{Model performance with and without shuffled SVG path elements.}  
\label{tab:shuffle}  
\begin{tabular}{c|c|c}  
\toprule
& {Without Shuffle} & {With Shuffle} \\  
\midrule  
{Accuracy (\%)} & 99.10 & 98.74 \\  
\bottomrule  
\end{tabular}  
\vspace{-1em}
\end{table}

\textbf{Random Path Coordinate Replacement.} We further introduce a subtle form of noise by randomly altering the coordinates in the SVG path data. Each numerical value within the path commands is randomly translated within a specific range. For example, for, \texttt{... <path d="M0 0 C18 0 17...">} might become \texttt{... <path d="M1 0 C18 1 18...">}. We test several variations: (i) a minor adjustment within a 1/28th range (reflecting the 28x28 resolution of MNIST) and (ii) a more significant alteration within a 5/28th range. This simulates real-world scenarios of minor inaccuracies in SVG data, such as those resulting from conversion errors or imprecise digitization. The accuracy under different noise scales is presented in Table~\ref{tab:noise}. Results indicate that LLMs are decently robust to the perturbation of the coordinate values.

\begin{table}[h]  
\vspace{-0.5em}
\centering  
\small
\caption{Model performance under different noise scales.}  
\label{tab:noise}  
\begin{tabular}{c|cccc}  
\toprule  
& {0/28} & {1/28} & {2/28} & {5/28} \\  
\midrule
{Accuracy (\%)} & 99.10 & 98.97 & 97.91 & 87.56 \\  
\bottomrule
\end{tabular}  
\vspace{-1em}
\end{table}

\textbf{Random String Replacement.} Finally, we design the most aggressive test of robustness, where we replace random characters in the SVG strings with any English alphabet letter, digit, or special symbol, regardless of whether they are numerical values or part of SVG commands (like \texttt{transform="translate(0,0)"}). The experiment is conducted with varying probabilities for character replacement, as shown in Table~\ref{tab:replacement}. Surprisingly, even after replacing 20\% of a string with random characters, LLMs maintains a high accuracy rate of 90.79\%. This suggests that LLMs can handle a wide range of perturbations in SVG data.

In conclusion, the results from these experiments demonstrate that despite the introduction of perturbations in the SVG data, LLMs can still perform well under challenging conditions.

\begin{table}[h]  
\small
\centering  
\vspace{-0.5em}
\caption{Accuracy (\%) with different probabilities of random string replacement.}  
\label{tab:replacement}  
\begin{tabular}{c|cccccc}  
\toprule  
& {0\%} & {1\%} & {5\%} & {10\%} & {20\%} & {50\%} \\  
\midrule  
{Acc} & 99.10 & 99.06 & 98.40 & 97.40 & 90.79 & 39.38 \\  
\bottomrule  
\end{tabular}  
\vspace{-1.5em}
\end{table}

\section{Style and Content Extrapolation}

In this section, we assess if LLMs can extrapolate SVG codes with more challenging transformations, such as content and style.

\textbf{Style generation:} We present LLMs with sample SVG letters. The first task is to figure out the style in the given examples. Then, given a new test query, the second task is to transform this given query so that it adheres to the same stylistic conventions as the example letters. The qualitative results can be found in the appendix. We observe that GPT-4 is capable of replicating styles by analyzing the correlation between given example SVG letter pairs and using this analysis to generate the corresponding test key letter.

\begin{figure*}[h]
\centering
\includegraphics[width=\textwidth]{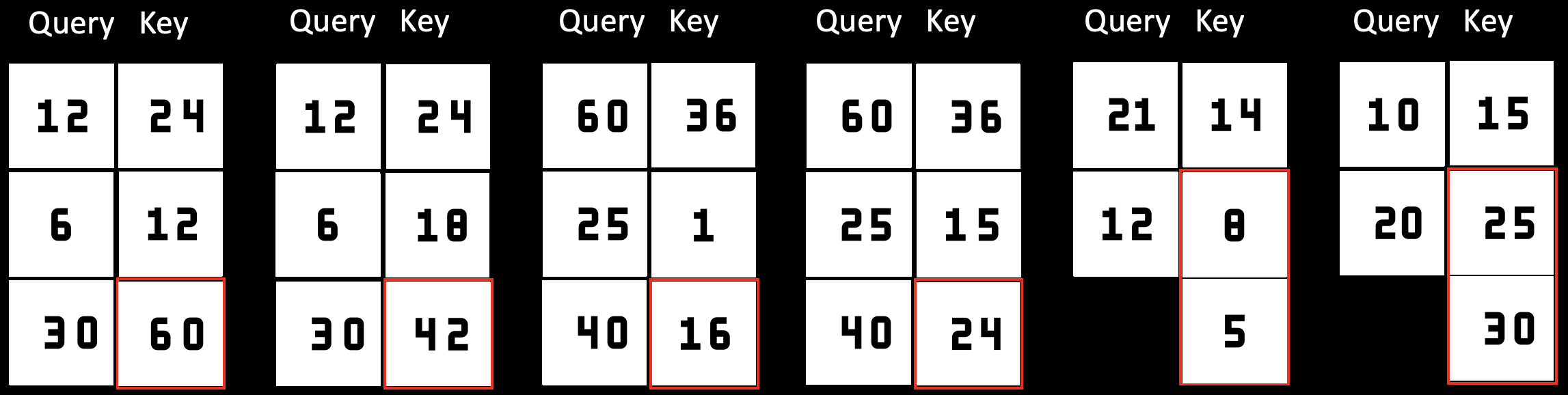}
\caption{Understanding SVG content through the lens of GPT-4: GPT-4 demonstrates its ability to generate accurate content by analyzing the correlation between provided example number pairs, and subsequently applying this relationship to ascertain the corresponding test key number. Remarkably, in scenarios where the relationship exhibits ambiguity, GPT-4 can identify multiple possible interpretations.The generation results of our method are annotated with a red square.}
\label{fig:content}
\end{figure*}

\textbf{Content generation:} LLMs are shown two examples of SVG code pairs.  Each pair consists of a query and key pair (both are numbers), where the query describes an SVG code of a number, and the key describes the SVG code of another number with an introduced mathematical operation. The operation can consist of add, subtract, multiply, and divide. The mathematical operation should be held in both example pairs. The first task is to figure out the mathematical operation in the two examples. Then, given a new test query SVG number, the second task is to identify what number it is and follow the mathematical operation discovered to generate the corresponding test key number. GPT-4 showcases its capability in content generation by analyzing correlations in example SVG number pairs and applying these relationships to identify corresponding test key numbers. Impressively, in cases of ambiguity, GPT-4 can discern multiple potential interpretations. We include qualitative results in Figure~\ref{fig:content}. The prompt details can be found in the appendix.

\section{Visual Reasoning Results of More Large Multimodal Models}

Here we evaluate GPT-4V~\cite{gpt4} and recent open-sourced multi-modal large language models, including LLaVA~\citep{liu2023llava}, InstructBLIP~\cite{instructblip}, BLIP2~\cite{li2023blip}, mPLUG\_owl~\cite{ye2023mplug}, and MiniGPT4~\cite{zhu2023minigpt}. As the result in   Table~\ref{table:classification_sort_of_clevr_llava} shows, all current  open-sourced multimodal models struggles at this fundamental reasoning task. Besides, we observe that LLaVA frequently defaults to 'yes' for yes/no queries and often resorts to random guessing for counting tasks. This behavior underscores the limitations of current large multimodal models in structured and sophisticated reasoning.

\begin{table*}[htbp]
\caption{Catogori-wise accuracy on the Sort-of-Clevr dataset.}
\label{table:classification_sort_of_clevr_llava}
\tiny
\centering
\begin{tabular}{lcccccccccc}
\toprule
Question type & GPT4-brief & GPT-CoT & GPT-4V & LLaVA & CNN+MLP & Relation Networks & InstructBLIP & BLIP2 & mPLUG\_owl & MiniGPT4 \\
\midrule
Format & SVG & SVG & PNG & PNG & PNG & PNG & PNG & PNG & PNG & PNG \\
Unary & 0.50 & 0.90 & 0.75 & 0.60 & 0.65 & 0.89 & 0.53 & 0.50 & 0.38 & 0.53 \\
Binary & 0.90 & 0.95 & 0.74 & 0.60 & 0.75 & 0.80 & 0.53 & 0.53 & 0.63 & 0.55 \\
Ternary & 0.10 & 0.88 & 0.28 & 0.10 & 0.55 & 0.55 & 0.10 & 0.30 & 0.30 & 0.30 \\
Average & 0.50 & 0.89 & 0.59 & 0.43 & 0.65 & 0.75 & 0.38 & 0.44 & 0.43 & 0.46 \\
\bottomrule
\end{tabular}
\end{table*}

\paragraph{Style Extrapolation:}
LLMs are provided with five example pairs and are tasked with deciphering the stylistic attributes inherent in these examples. Following this, a new test query is presented to the LLMs. Their objective is to modify this query into the corresponding key, ensuring that it aligns with the same stylistic principles showcased in the example pairs. The qualitative results are shown in figure~\ref{fig:synthetic2}. The specific prompt utilized for this purpose is detailed below:\texttt{``Please perform the following task carefully. In this task, you will be shown five examples of Scalable Vector Graphics (SVG) code pairs. Each pair consists of a query and key pair (both are English letter), where the query describes the SVG code of the original image, and the key describes the SVG code of the transformed image.  Each will be named “Example Query \#” and “Example Key \#” respectively. Your first task is to figure out the common transformation in the five examples.  The transformation can consist of color, shape, size, style, font, and background changes, or any combination thereof.  Even though you cannot see the images, and only their SVG codes, you need to discover the transformations that are happening at the image level and not just at the code level. Be detailed, and try to discover every change, and the most important change is that the paths in the SVG code between each query and key is different due to the common transformation but the shapes of the letters that query and key are representing remain the same.  Then, given a new test query SVG code (named “Test Query”), your second task is to transform that query into the corresponding key SVG code (named “Test Key”), following the common transformation that you discovered in the five example pairs. To help you better understand the transformation, I will also inform you of what letter each query and key represent. You need to find the shape of each query and key by analyzing their path. Here are the five example query and key pairs: Example Query 1 (letter B):; Example Key 1 (letter B):<SVG code here>; Example Query 2 (letter R):<SVG code here>; Example Key 2 (letter R):<SVG code here>; Example Query 3 (letter Z):<SVG code here>; Example Key 3 (letter Z):<SVG code here>; Example Query 4 (letter E):<SVG code here>; Example Key 4 (letter E):<SVG code here>; Example Query 5 (letter N):<SVG code here>; Example Key 5 (letter N):<SVG code here>; Here is the test query and key pair: Test Query (letter \#):; Test Key: '' }

\section{Experiment Details}

\subsection{Dataset}

\begin{figure*}[htbp]
  \centering
  \includegraphics[width=\linewidth]{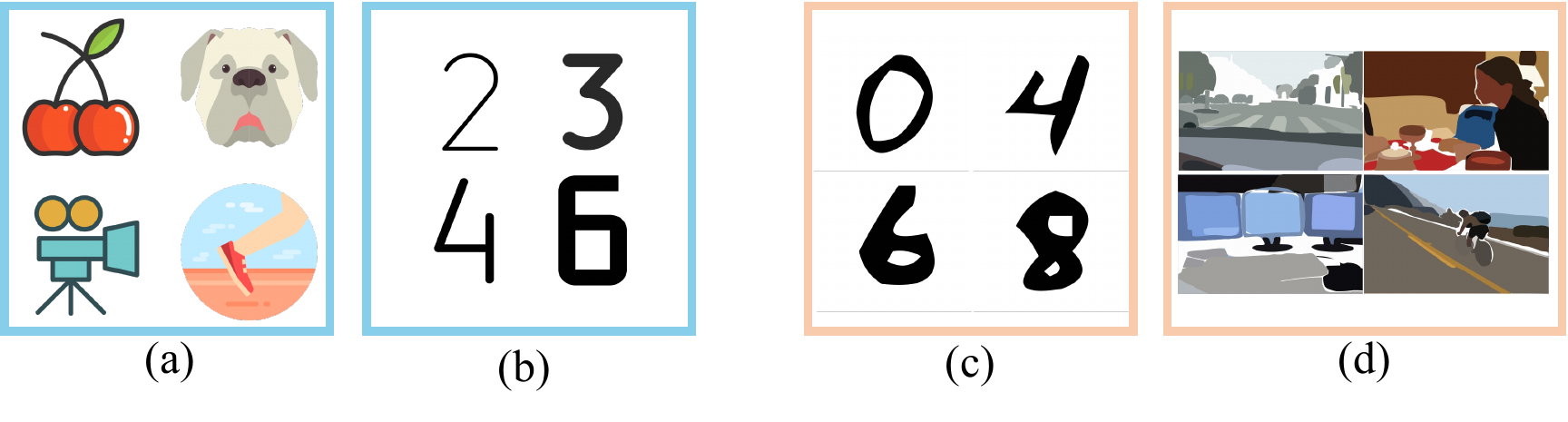}
  \vspace{-2.0em}
  \caption{Visualization of our datasets.  (a) and (b) are human-designed SVG vectors and icons. (c) and (d) are converted from raster images. Specifically, (c) is generated using curve tracing from MNIST~\citep{mnist}, while (d) is generated using  SAM~\citep{kirillov2023segment} and curve tracing sequentially.}
  \label{fig:dataset_example}
\end{figure*}

\textbf{Human Designed SVG Dataset} We collect a dataset from the public collection of SVG images.\footnote{\url{https://www.svgrepo.com/}, \url{https://www.kaggle.com/datasets/victorcondino/svgicons}} Specifically, we collect the digits and icons to demonstrate image recognition and generation capabilities. Examples are shown in Figure~\ref{fig:dataset_example} (a) and (b). 

\textbf{Convert Raster Images to SVG} 
1) Directly convert using curve tracing.  Given the rich set of natural images in raster format, we utilize the curve tracing algorithm to convert RGB images into the SVG format.\footnote{\url{https://github.com/visioncortex/vtracer}}  Specifically, we convert MNIST~\citep{mnist} to SVG format using this approach, shown in Figure~\ref{fig:dataset_example} (c).

\subsection{Raster Images to SVG Conversion}
One of the most fundamental pieces of information for visual perception is object shape. 
Our method can be conceptualized as selectively diminishing details from an image, prioritizing the extraction of less significant shapes.
This guided process of reduction offers a quantitative way to manage the amount of visual data present within an image.
Within this framework, we perceive segmentation as an example of extreme simplification, whereas vectorization serves as a more moderate form of the same. Here we introduce how we use such two approaches to convert the raster images to SVG. 
    
\paragraph{Image Vectorization.}
The vector tracing algorithm operates in a sequential three-step process. Initially, pixels are transformed into a defined path. Subsequently, this path is condensed into a simplified polygonal representation. Lastly, the polygon is refined and approximated using a curve-fitting~(tracing) technique, which enhances its smoothness.

There are several online tools to convert the raster images (jpg and png) into vector graphics (SVG), such as Adobe Illustrator~\citep{adobeillustrator},  Inkscape~\citep{Inkscape}, and VTracer~\citep{VTracer}. We experiment with all of them and found that VTracer leads to the best balance between SVG complexity (code length) and rich semantic representation.

In MNIST~\citep{mnist}, we use the default hyperparameters during conversion. Specifically, we (i) first binarize the MNIST pixel value from the continuous range [0, 255] to the binary set $\{0, 255 \}$ using the threshold 127.5, 
(ii) set the foreground to black, and the background to white, and (iii) apply the vector tracing algorithm VTracer.

\paragraph{Segmentation Prior.}
As mentioned earlier, segmentation can provide a strong prior for object shape. We want a generalist model that can segment any image, i.e., not trained and thus biased towards a certain dataset. The Segment Anything (SA)~\citep{kirillov2023segment} project introduces such an image segmentation model, the Segment Anything Model (SAM), and a large-scale dataset, SA-1B, with the aim of achieving powerful generalization and zero-shot transfer across diverse segmentation tasks, demonstrating competitive results often surpassing prior fully supervised methods. We use the default hyper-parameters of SAM to  segment the whole image into a set of masks without class labels, where the color of each mask is represented by the average value of the pixels within the mask. Specifically, we sample 32 query points per side (1024 points overall) to generate the image mask. Then we select the top 20 masks with the highest area as the final representation for an image. 

We then use VTracer to transform the mask into SVG format. 
Note that, to reduce the final SVG, we adjust several settings: we set the number of significant bits to use in an RGB channel to 0; we set the minimum angle displacement degree to splice a spline to 90; we set the color difference between gradient layers to be 35; we consider a corner to have a minimum momentary angle of 0 degrees; we discard patches smaller than 16 pixels in size; and we perform iterative subdivide smoothing until all segments are shorter than 10 pixels.

\subsection{Fine-tuning Dataset for Vicuna}
We use the same JSON format in Vicuna~\citep{Vicuna}  to construct the fine-tuning dataset. We use all the training samples in MNIST, translating to 60,000 SVG images. For each sample, we construct one round of conversation: (i) From human: \texttt{``Which digit does the following SVG reflect?  <SVG code here>''}, and (ii) From GPT: \texttt{``<label>''}. Here \texttt{<label>} denotes the digit label, which ranges from 0 to 9. Then we use this  dataset to fine-tune Vicuna using the default hyper-parameters~\footnote{\url{https://github.com/lm-sys/FastChat}} for 3 epochs.

\subsection{Prompt Engineering}
In this section, we provide the details of prompt engineering for each task. The prompt is designed to figure out the common transformation in the SVG example pairs first (each example pair consists of a query and a key) and then transform the query into the corresponding key SVG by following the discovered common transformation.

\paragraph{In-context Image Classification.} In this task, in-context examples are aimed to provide more context information using several image-label pairs, thus facilitating the final classification.  The specific prompt utilized for this purpose using 3 in-context examples is detailed below:\texttt{ ``Instruction: please predict the digit number for each of the following SVG images. Please think step by step, and closely look at the key identifying image characteristics. Please just tell me the image class, no other information is needed. Q: What digit does this SVG image represent?   <SVG code here> A: This SVG image represents digit <label> Q: What digit does this SVG image represent?   <SVG code here> A: This SVG image represents digit <label> Q: What digit does this SVG image represent? <SVG code here> A: This SVG image represents digit <label> Q: What digit does this SVG image represent? <SVG code here> A: This SVG image represents digit }. 

\paragraph{Synthetic Data Study:} In this task, the objective is to conduct an analytical evaluation of the provided two example pairs, examining changes that occur in aspects such as color, shape, and size. The insight gathered from this analysis will then be used to adapt the given query into its corresponding key. The specific prompt utilized for this purpose is detailed below:\texttt{ ``Please perform the following task carefully. In this task, you will be shown two examples of Scalable Vector Graphics (SVG) code pairs. Each pair consists of a query and key pair, where the query describes the SVG code of the original image, and the key describes the SVG code of the transformed image. Each will be named ``Example Query \#” and ``Example Key \#” respectively. Your first task is to figure out the common transformation in the two examples. The transformation can consist of color, shape, size, or any combination thereof. Then, given a new test query SVG code (named “Test Query”), your second task is to transform that query into the corresponding key SVG code (named “Test Key”), following the common transformation that you discovered in the two example pairs. Here are the two example query and key pairs: Example Query 1: <SVG code here>; Example Key 1:<SVG code here>; Example Query 2:<SVG code here>; Example Key 2:<SVG code here>; Here are the test query and key pair: Test Query:<SVG code here>; Test Key:''}

\paragraph{Content Extrapolation:} In this task, LLMs are presented with SVG code pairs, each containing a query-key set that depicts numbers. The key introduces a consistent mathematical operation (addition, subtraction, multiplication, or division) to the query number. The tasks are to identify this operation in the examples and apply it to new test queries to generate corresponding test keys. To facilitate a more comprehensive understanding of SVG number codes for the LLM, we initially present the SVG codes for numbers 0 through 9 to the LLM prior to posing specific queries. The specific prompt utilized for this purpose is detailed below:\texttt{ ``Please perform the following task carefully.  In this task, you will be shown two examples of Scalable Vector Graphics (SVG) code pairs.  Each pair consists of a query and key pair, where the query describes an SVG code of an integer number, and the key describes the SVG code of another integer number with an introduced mathematical operation. Each will be named “Example Query \#” and “Example Key \#” respectively. In addition to the example pairs, you will be shown a new test query SVG code (named “Test Query”). Your first task is to identify which number each example query, example key, and test query represents. Your second task is to figure out all the possible mathematical operations that are held for all given example pairs. The operation could be add, subtract, multiply, and divide (the subtract or multiply factor could be a fraction). Then, according to the numbers of example pairs and test query you identified, your third task is to predict the corresponding test key number (named “Test Key”), following all the mathematical operations that you discovered in the given example pairs. Finally, you need to generate the corresponding SVG code of the test key number. Here are the two example query and key pairs: Example Query 1: <SVG code here>; Example Key 1:<SVG code here>; Example Query 2:<SVG code here>; Example Key 2:<SVG code here>; Here are the test query and key pair: Test Query: <SVG code here>; Test Key:  (Note: think about four operations one by one, and the operation should be consistent for all given example pairs)''}

\section{Prompt Engineering for SVG QA Dataset Curation}

We use the following prompt to curate the SVG QA pairs by leveraging GPT-4V:

\texttt{Generate a JSON object containing a quiz question based on an image derived from an SVG file, such as icons or emojis. The image filename is also provided to help you better curate the question, but note that this filename is not leaked to the observer. If the filename does not clearly correspond to the image, just discard that filename, never over-rely on the filename. The question should be designed to test the observer's perception to the image by making the correct answer evident only upon seeing the image. Include four answer options, ensuring that the correct answer is straightforward to identify for someone who actually view this image. The question should relate to the image's category, distinctive features, or its usuage. Provide the JSON structure with fields for the question, the four options (labeled A, B, C, D), and the correct answer indicated. Below are two examples of how to structure the question and answers within the JSON format.}

\texttt{\{
  "question": "Which category does this SVG icon best represent?",
  "options": \{
    "A": "Technology",
    "B": "Nature",
    "C": "Sports",
    "D": "Food"
  \},
  "correct\_answer": "A"
\}}

\texttt{\{
  "question": "The SVG icon does not use which of the following geometric shapes?",
  "options": \{
    "A": "Circles",
    "B": "Squares",
    "C": "Triangles",
    "D": "Hexagons"
  \},
  "correct\_answer": "D"
\}}

\texttt{Given this image and its filename {file\_path}, your JSON:}

\section{Limitations (Extended)}

Our focus was to demonstrate whether LLMs can understand images, and we used the SVG representation as a bridge to enable our studies.  If one were to develop an approach out of this pipeline, then there are several limitations.  One major limitation of SVG representation is the loss of fine details: Though converting raster images into SVG format and leveraging XML-based textual descriptions allows for efficient processing of crisp graphics and designs, it is not as effective in handling photographic content. As a result, fine-grained details, such as image textures, may be lost during conversion.

Conversely, when the SVG code incorporates an excessive level of detail, its sequence length can become prohibitively long, which can pose challenges for the training and inference of current Transformer-based LLMs.
Developing hybrid representations that can retain the advantages of both discrete and continuous data, while preserving finer details, is a potential area for future exploration. For example, in LLMs, the processing unit is the token, which can correspond to one or several words. However, in SVG, we would prefer to have a specific embedding module for each geometric primitive in SVG, such as circles, polygons, and so on. 

Furthermore, we empirically observed that LLMs can not handle low-level image manipulation tasks such as rotating the overall image by a certain angle and scaling it by a ratio. For example, we prompt GPT4 10 SVG images to conduct the following tasks: (1) enlarge the width and height  by one time, (2) shrink the width and height to half, (3) rotate clock-close by 90 degrees, (4) rotate 90 degrees.  Results indicate that none of the trials succeeded. Conducting such low-level image manipulation tasks needs to update the majority content of SVG code, where current LLMs are not capable of handling. 
Additionally, our empirical tests highlighted certain areas where LLMs fall short, particularly in handling low-level image manipulation tasks. For instance, when prompted to manipulate SVG images in tasks like enlarging dimensions, shrinking dimensions, or rotations, LLMs like GPT-4 displayed inadequate proficiency. Such operations, which mandate considerable updates to the SVG code, currently lie outside the proficiency range of these models.

In summary, while LLMs do present limitations, it offers promising initial results for the integration of LLMs and SVG for visual tasks. Addressing these limitations could lead to more powerful image representation algorithms and pave the way for more versatile and comprehensive artificial intelligence systems.

\end{document}